\pdfoutput=1

\documentclass[11pt]{article}

\usepackage[preprint]{acl}

\usepackage{times}
\usepackage{latexsym}

\usepackage[T1]{fontenc}

\usepackage[utf8]{inputenc}

\usepackage{microtype}

\usepackage{inconsolata}

\usepackage{graphicx}

\usepackage[capitalize,nameinlink,noabbrev]{cleveref}

\usepackage{enumitem}

\usepackage[title]{appendix}

\definecolor{cb-black}      {RGB}{  0,   0,   0}
\definecolor{cb-blue-green} {RGB}{  0,  073,  073}
\definecolor{cb-green-sea}  {RGB}{  0, 146, 146}
\definecolor{cb-rose}       {RGB}{255, 109, 182}
\definecolor{cb-salmon-pink}{RGB}{255, 182, 119}
\definecolor{cb-purple}     {RGB}{ 73,   0, 146}
\definecolor{cb-blue}       {RGB}{ 0, 109, 219}
\definecolor{cb-lilac}      {RGB}{182, 109, 255}
\definecolor{cb-blue-sky}   {RGB}{109, 182, 255}
\definecolor{cb-blue-light} {RGB}{182, 219, 255}
\definecolor{cb-burgundy}   {RGB}{146,   0,   0}
\definecolor{cb-brown}      {RGB}{146,  73,   0}
\definecolor{cb-clay}       {RGB}{219, 209,   0}
\definecolor{cb-green-lime} {RGB}{ 36, 255,  36}
\definecolor{cb-yellow}     {RGB}{255, 255, 109}

\newcommand{\cbGreenSea}   [1]{\textcolor{cb-green-sea}  {#1}}
\newcommand{\cbRose}       [1]{\textcolor{cb-rose}       {#1}}

\newcommand{\cbBlue}       [1]{\textcolor{cb-blue}       {#1}}
\newcommand{\cbLilac}      [1]{\textcolor{cb-lilac}      {#1}}

\newcommand{\cbBurgundy}   [1]{\textcolor{cb-burgundy}   {#1}}

\title{Sleepless Nights, Sugary Days:\\Creating Synthetic Users with Health Conditions \\for Realistic Coaching Agent Interactions}

\author{
\textbf{
 Taedong Yun\textsuperscript{1,\textdagger},
 Eric Yang\textsuperscript{2},
 Mustafa Safdari\textsuperscript{1},
 Jong Ha Lee\textsuperscript{2},
}
\\
\textbf{
 Vaishnavi Vinod Kumar\textsuperscript{3},
 S. Sara Mahdavi\textsuperscript{1},
 Jonathan Amar\textsuperscript{2},
 Derek Peyton\textsuperscript{3},
}
\\
\textbf{
 Reut Aharony\textsuperscript{1},
 Andreas Michaelides\textsuperscript{3},
 Logan Schneider\textsuperscript{3},
 Isaac Galatzer-Levy\textsuperscript{3},
}
\\
\textbf{
 Yugang Jia\textsuperscript{2},
 John Canny\textsuperscript{1},
 Arthur Gretton\textsuperscript{1,*,\textdagger},
 Maja Matari\'c\textsuperscript{1,*,\textdagger}
}
\\
 \textsuperscript{1}Google DeepMind \phantom{ }
 \textsuperscript{2}Verily Life Sciences \phantom{ }
 \textsuperscript{3}Google
\\
 \small{\textsuperscript{*}Co-last authors (listed alphabetically)}
\\
 \small{
 \textsuperscript{\textdagger}\texttt{\{tedyun,gretton,majamataric\}@google.com}
 }
}

\begin{document}
\maketitle

\newcommand{\cmt}[1]{\textcolor{red}{[#1]}}  

\begin{abstract}
We present an end-to-end framework for generating synthetic users for evaluating interactive agents designed to encourage positive behavior changes, such as in health and lifestyle coaching. The synthetic users are grounded in health and lifestyle conditions, specifically sleep and diabetes management in this study, to ensure realistic interactions with the health coaching agent. Synthetic users are created in two stages: first, structured data are generated grounded in real-world health and lifestyle factors in addition to basic demographics and behavioral attributes; second, full profiles of the synthetic users are developed conditioned on the structured data. Interactions between synthetic users and the coaching agent are simulated using generative agent-based models such as Concordia, or directly by prompting a language model. Using two independently-developed agents for sleep and diabetes coaching as case studies, the validity of this framework is demonstrated by analyzing the coaching agent's understanding of the synthetic users' needs and challenges. Finally, through multiple blinded evaluations of user-coach interactions by human experts, we demonstrate that our synthetic users with health and behavioral attributes more accurately portray real human users with the same attributes, compared to generic synthetic users not grounded in such attributes. The proposed framework lays the foundation for efficient development of conversational agents through extensive, realistic, and grounded simulated interactions.
\end{abstract}

\section{Introduction}
\label{sec:intro}

Personalization is a key capability for any interactive agents aiming to encourage positive behavior changes, for instance in health and lifestyle \citep{christakopoulou2024agentsthinkingfastslow,yang2024barrierstacticsbehavioralscienceinformed}. The efficacy of such agents must be assessed via interactions with users; however, collecting and evaluating diverse, long-term human interactions with agents is costly and time-consuming, and the development of such agents can be significantly accelerated with realistic simulation. We address the challenge of realistic user simulation in this context by proposing an end-to-end framework for generating cohorts of synthetic users grounded in health, lifestyle, and behavior, and having them interact with coaching agents. The framework is generally applicable across a range of health and lifestyle domains; we validate it in the domains of sleep and diabetes coaching.

Synthetic users generated with large language models (LLMs) have been widely studied in recent years \citep{kapania2024simulacrumstoriesexamininglarge,wang2025opencharactertrainingcustomizableroleplaying,moon2024virtual,park2024generativeagentsimulations1000,joonsung_simulacra}, often for general purpose alignment and personalization \citep{castricato2024personareproducibletestbedpluralistic}, and have been designed to reflect demographic information about the population at large. 
We focus on a more targeted application, where the synthetic users are employed for interaction with foundation models or agents specialized in health, as done in recent work  \cite{tu2024conversationaldiagnosticai,yu2024aipatientsimulatingpatientsehrs,Johri2025-kp}. We provide an overview of such work in \cref{sec:background}.

We demonstrate an end-to-end-framework for designing a cohort of synthetic users exhibiting health conditions consistent with their specific demographics and behavioral factors, in order for their interaction with a coaching agent to realistically reflect their needs and challenges. We construct the synthetic users by first generating a sample of natural language "vignettes", which are in turn generated based on real demographics, health, and behavioral data  \citep[such as the Big Five markers;][]{goldberg1992development,serapiogarcia2023personalitytraitslargelanguage}. Our framework also allows for optionally including additional information:  domain-specific rubrics such as ``barriers'' and ``goals'' as used by human coaches; challenges to action derived from the COM-B behavioural model \citep{Michie2011-gp}; or rich backstories for the synthetic users, generated consistently with the structured attributes from real data.
Finally, based on this cohort of vignettes, realistic user-coach simulation can be performed by direct LLM calls or by using generative agent-based models such as the Concordia \citep{Vezhnevets2023-vq} open-source system. More details will be discussed in \cref{sec:methods}.

We demonstrate the efficacy of our methodology for generating health-grounded synthetic personas in two separate domains, sleep coaching (\cref{sec:sleepCoaching}) and diabetes coaching (\cref{sec:diabetesCoaching}), using two independently developed sleep coaching and diabetes coaching agents. We verify that our synthetic users indeed communicate behaviors, health conditions, and barriers consistent with their assigned attributes, by inspecting the internal state-of-user model of the coaching agent for each synthetic user, by a holistic evaluation of the user-coach interactions by trained human experts, and by a comparative evaluation of our health-grounded synthetic users against generic synthetic users, again by trained human experts.

\section{Background}
\label{sec:background}

In this section, we briefly survey the development and goals of LLM-based synthetic personas in general, and then provide further details on prior literature on the design and application of synthetic personas for health. We next describe software systems for simulating agent interactions. Finally, we discuss some challenges in using LLM-derived synthetic users for agent development and beyond.

\subsection{LLMs as Synthetic Personas}

The time and expense required in human evaluation of LLM agents naturally evokes the need for auto-evaluation. This has led to the development of ``reinforcement learning from AI feedback (RLAIF)'', where LLMs serve to evaluate the output of other LLMs, in  a now  well-established strategy for maximizing alignment, factuality, helpfulness, and other desiderata  \citep{bai2022constitutionalaiharmlessnessai,lee2024rlaifvsrlhfscaling}.

One recent approach to RLAIF uses  LLMs to construct synthetic personas and to judge their interaction with the model being evaluated.
\citet{castricato2024personareproducibletestbedpluralistic} generated these synthetic personas 
from the US census data, and employed them to build an evaluation dataset of prompt and feedback pairs obtained solely from synthetic users. A strength of the approach is that results for specific users can be modeled, rather than general user categories. This approach also motivates the work of \citet{moon2024virtual}.

In addition to demographic information, \citet{serapiogarcia2023personalitytraitslargelanguage} showed that LLMs can be designed to display specific personality traits, such as those along the Big Five dimensions \citep{goldberg1992development}.
Effective modeling and control of personality in LLMs is important for two reasons: first
since personality is a key factor in determining effective communication; second, because it is important to model interaction of synthetic users over the broad range of personality profiles to be encountered in practice. \citet{serapiogarcia2023personalitytraitslargelanguage} found that instruction-tuned models displayed more reliable and externally valid personality types 
than pre-trained variants (perhaps due to superior instruction following), and that
larger models are better equipped to express complex traits, and to simulate social behaviours. 



\subsection{LLMs and Synthetic Personas in Health}

Several works have proposed the use of LLMs to provide advice in health or wellness settings.

\citet{tu2024conversationaldiagnosticai} introduced Articulate Medical Intelligence Explorer (AMIE),  an LLM agent designed for conversational medical diagnosis. The AMIE agent achieved remarkable success, with greater diagnostic accuracy compared to primary care physicians in a randomized, double-blind crossover study.
AMIE was trained using self-play with simulated doctor-patient dialogues and was tested with patients simulated first by searching for medical conditions from across three databases (Health QA, MalaCards Human Disease, MedicineNet Diseases \& Conditions), and then by performing internet searches on each medical condition in order to retrieve passages on demographics, symptoms, and management, which were screened for relevance. PaLM 2 \citep{anil2023palm2technicalreport} was used as the base model for all agents, including the patients and doctors, which were given role-specific prompts.
Instruction fine tuning of both patient and doctor took place, first using static datasets of real doctor-patient interactions, and subsequently using dialogues generated through self-play. A moderator agent is used to manage the dialogue between doctor and patient, and to decide on when the conversation has ended. 

There were a number of limitations to the synthetic user (patient) design in the AMIE study. First, the retrieval of symptoms and demographics from online discussion might create bias towards demographics of individuals most likely to engage in online discussion of their condition. Second, online discussion might misattribute irrelevant symptoms to conditions, and be misinformed as to mitigation strategies. Third, the model did not explicitly control for personality traits, relying on the distribution of personality traits to be found in online discussion of the medical conditions, which might not be representative of target demographics. Related to this issue, PaLM 2 has undergone a fine-tuning process which may make the patient agents more obliging, sympathetic, and communicative than real patients.

\citet{yu2024aipatientsimulatingpatientsehrs} also described an application of LLMs to simulate user (patient) populations, based on particular medical histories from 1495 patients in the MIMIC-III database \citep{Johnson2016-eg}. The approach represented each patient's medical and demographic status as a knowledge graph (with 15,000 possible nodes and over 26,000 possible edges), then retrieved patient information from this graph in response to doctor queries. The retrieved answers were then expressed in natural language, with tone adapted in accordance with the Big Five personality traits. The key finding was that ``patient LLMs'' that retrieve from a structured representation of patient traits more accurately communicate symptoms and history than simply using unstructured clinical notes in the prompt. 

This past work differs from our approach in a number of important ways. First, in the past work, the representation of each individual as a knowledge graph is suited to clinical settings, where unstructured clinical notes must be organized as a pre-processing step. Accordingly, much of the focus of their work is on the construction of the graph, and accurate retrieval from it. The graph representation may be less suited to wellness settings, however, where the user's needs and challenges require less structure and granularity.  Second, and more importantly, the generation process can only traverse the graph of existing patients, and cannot generate new individuals. This may lead to limitations with respect to both privacy and scalability.

\citet{Johri2025-kp} described the use of synthetic personas in evaluating an LLM system for patient history collection in medicine. The patient agent was instructed explicitly not to display medical knowledge or generate new symptoms beyond what was given in the ``vignette''.
A total of 2,000 case vignettes were considered: 1,800 questions from MedQA-United States Medical Licensing Examination covering 12 medical specialties, and 200 questions on skin disease from the Derm-Public and Derm-Private question banks.
While representing a broad range of conditions, the focus of that study was primarily on evaluating information gathering and diagnosis abilities of LLMs, rather than coaching. Unlike our work, \citeauthor{Johri2025-kp} did not attempt to ensure that patient profiles were representative of the population at large, nor were personality types modeled when generating patient dialogues.

\subsection{Health Coaching Datasets}

There are a limited number of datasets on patient conversations with health practitioners, such as \citet{zhou-etal-2024-modeling} and \citet{wang-etal-2023-medngage}.

In brief, the \citet{zhou-etal-2024-modeling} dataset consists of a cohort of 22 patients for up to eight weeks, and records both the initial S.M.A.R.T. (specific, measurable, achievable, relevant, and time-bound) goal creation, and subsequent progress monitoring and engagement. \citet{zhou-etal-2024-modeling} focus on a neurosymbolic representation of goals negotiated in collaboration with a coach, which can include very broad wellness categories. On the other hand, our focus is on more sharply defined wellness issues (sleep or diet) where specific goals and challenges are addressed. We also address only the elicitation of barriers and goals, not subsequent behaviour change. 

\citet{wang-etal-2023-medngage} provides a dataset of 68 patient dialogues in the setting of engagement with medical practitioners for managing cancer symptoms. Their use of LLMs to monitor engagement is a promising direction for future updates to the coaching agent in our study.

\begin{figure*}[t]
  \centering
  \includegraphics[width=\textwidth]{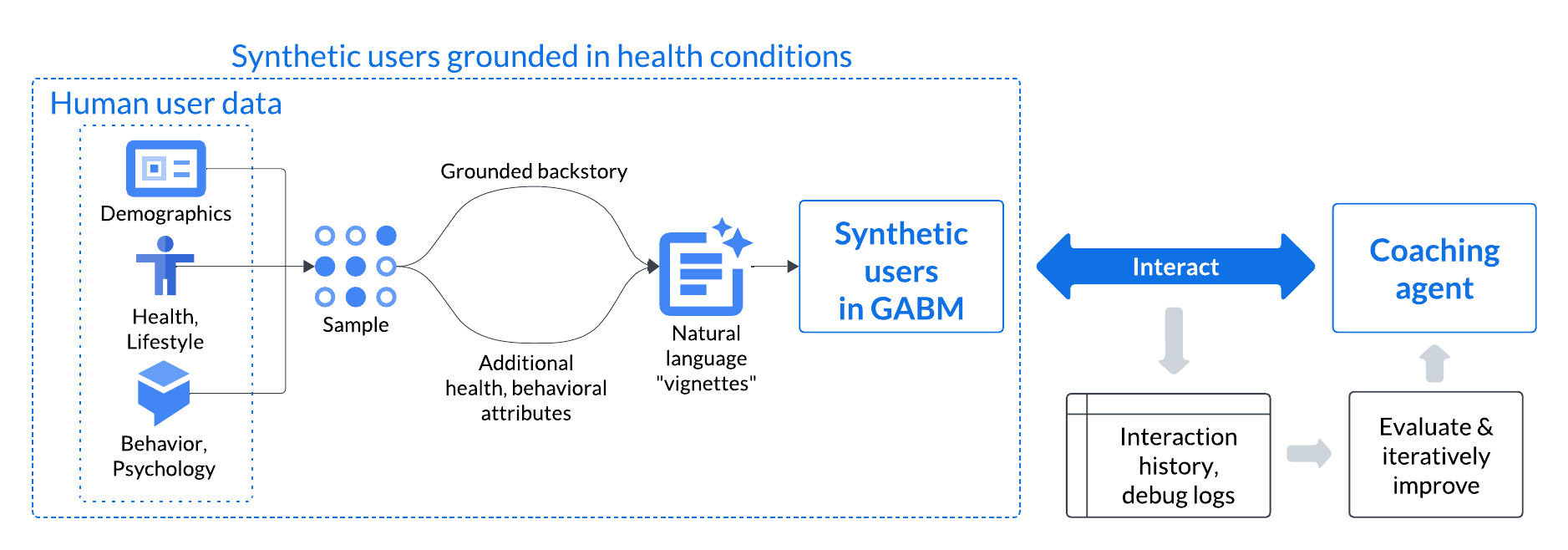}
  \caption{An overview of generating synthetic users grounded in real demographics, health \& lifestyle, and behavioral \& psychological characteristics for automated evaluation of coaching agent interactions. GABM: generative agent-based model.}
  \label{fig:overview}
  \vspace{-1.4em}
\end{figure*}

\subsection{Generative Agent-Based Models}
\label{background:concordia}
Generative agent-based models, such as the Concordia system which we used \citep{Vezhnevets2023-vq}, are directly relevant to user persona simulation. Concordia offers a framework for generating agent interactions, particularly dialog-based ones, given a sufficient backstory. Its design incorporates several key features beneficial for research in generative agent-based simulation. First, Concordia employs associative memory \cite{joonsung_simulacra} and chain-of-thought reasoning \cite{wei_reasoning2022} to ensure that agent utterances are grounded in the provided backstory. Second, it utilizes metadata tags within prompts to clearly delineate different contextual elements, including agent memory, observations, prior conversation turns, and overarching goals. Furthermore, these contextual components can be assigned varying levels of importance, guiding the underlying LLM to generate subsequent dialog using appropriate chain-of-thought reasoning. Third, Concordia provides extensive logging capabilities, enabling detailed examination of the pipeline's internal processes, such as the constructed prompts sent to the LLM, the chain-of-thought reasoning process, post-processing steps, and the final output. Fourth, its modular architecture allows for flexible configuration, enabling researchers to customize the agent pipeline by swapping components.

A crucial requirement for generative agent-based simulation frameworks is the ability to integrate and manage multiple, mutable states within the interaction environment, as these states can significantly influence an agent's subsequent actions and utterances.  Critically, the mutation of these states, often lacking a true real-world simulation, must also be driven by chain-of-thought reasoning. Concordia addresses this need by providing code abstractions that facilitate the integration and management of these dynamic environmental states and their reasoned evolution.

\subsection{Challenges of Using LLM-Derived Synthetic Personas}

As discussed by \citet{kapania2024simulacrumstoriesexamininglarge}, there are a number of challenges in using LLMs as synthetic personas (users). LLMs are a consolidation over human experience, and hence they combine multiple viewpoints, failing to represent the diversity of human experience or correlations between different aspects of individuals. They can also lack depth: for instance, they can make references to particular phenomena (e.g., sleep difficulties in our test domain), but these phenomena do not represent situated knowledge, grounded in lived experience and history. This makes it difficult to model human users, since the relevant reasons for the health conditions might be missing. We mitigate this issue with the option of adding backstories, which have been shown to provide grounding to LLM-generated output \citep{moon2024virtual}. 

Training data may also be biased as they were drawn largely from English-speaking cultures and individuals with strong online presence. To demonstrate the issue in the domain of sleep disorders,  a survey of 27,000 US adults revealed that difficulty in falling and in staying asleep occurs more commonly among residents in non-metropolitan areas, and for those on lower incomes \citep{Adjaye2022sleepDifficulties}. We address this bias by specifying persona statistics drawn from relevant demographic health studies  \citep[e.g.,][]{Yfantidou2022-ay,Arges2020-zh}, rather than relying on generation by the LLM. 



A further subtle risk of using LLMs in simulating synthetic users can be understood from a causal inference perspective, as discussed in \citet{Gui_2023}. Providing particular advice to an LLM (e.g., a recommendation regarding better sleep practices) can inadvertently cause unintended variation in other factors  (e.g., the LLM's beliefs about the age and lifestyle factors of the synthetic persona). We mitigate this effect through explicit control over demographic and personality factors that the LLM might otherwise incorrectly impute, a practice recommended by \citet{Gui_2023}, although this remains an ongoing research challenge. 

\section{Methods}\label{sec:methods}

While LLMs trained on a large data corpus (e.g., the entire Internet) partially represent the general human population, the distributions of subpopulations and health conditions represented in a single model are skewed and poorly understood \citep{kapania2024simulacrumstoriesexamininglarge}. Furthermore, the instruction fine-tuning step (e.g., reinforcement learning with human feedback) typically employed by high-performance LLMs further impacts the distribution of generated outputs. In this work we propose an end-to-end framework for generating synthetic users grounded in reality, by starting from human user data including demographics, health \& lifestyle, and behavioral \& psychological characteristics, to accurately represent a desired population (\cref{fig:overview}). This real cohort can be  sampled either uniformly, or with oversampling or undersampling explicitly designed by researchers (e.g., to better surface underrepresented subgroups or conditions). Given this sampled cohort, we optionally add additional health conditions conditioned on already specified attributes of the synthetic users, either by using the LLM or via a rule-based algorithm, and also optionally add rich backstories of the synthetic user, conditioned on existing attributes \citep{moon2024virtual}.

This combination of background information grounded in real data constitutes a ``vignette'' of the synthetic user, all based in natural language. In the following sections we explore how these vignettes can be used to generate interactions between synthetic users and coaching agents, in two separate  health coaching scenarios--sleep coaching and diabetes coaching--with two independently developed coaching agents.

\section{Experiments: Sleep Coaching}\label{sec:sleepCoaching}

Sleep is vital for health and well-being; insufficient sleep and untreated sleep disorders can have a detrimental effect on cognitive function, mood, mental health, and cardiovascular, cerebrovascular, and metabolic health \citep{Ramar2021-ql}.
Sleep disorders are highly prevalent: according to the 2020 National Health Interview Survey (NHIS) \cite{Adjaye2022sleepDifficulties},  14.5\% of US adults had trouble falling asleep most days over the 30-day study period, and over a quarter of adults do not meet the minimum recommended sleep duration per night.

\subsection{Synthetic Users with Sleep Conditions}

To generate synthetic users grounded in real sleep conditions, we utilized the publicly available LifeSnaps dataset, a multi-modal, longitudinal, geographically-distributed dataset containing participant demographics, smartwatch measurements including sleep data, health, behavioral, and psychological trait surveys, and ecological momentary assessments \citep{Yfantidou2022-ay}. Among the fields available in this dataset, we focused on basic demographics (age, gender), basic health \& sleep attributes (body mass index, sleep duration \& efficiency), and five personality markers from the International Personality Item Pool (IPIP) version of the Big Five \citep{goldberg1992development}. Importantly, given  longitudinal sleep data over four months in LifeSnaps, we included both the average sleep duration and the variability of sleep duration, both important health factors. See \cref{appx:lifesnaps_detail} for more details.

As outlined in \cref{fig:overview}, we generated a synthetic ``sleep profile'' for each individuals in LifeSnaps ($N$=68) conditioned on real demographics, sleep duration \& efficiency, and behavioral characteristics. The generated sleep profile consists of the following four key attributes: ``primary sleep concern'', ``sleep goals'', ``reasons for goals'', and ``barriers''. The full prompt used to generate these attributes is available in \cref{appx:sleep_profile}. This sleep profile and all aforementioned structured attributes from LifeSnaps formed the ``vigenettes'' for our synthetic users with sleep conditions.

Finally, we instantiated each synthetic user in Concordia \citep{Vezhnevets2023-vq} as a ``SimpleLLMAgent'' to prioritize clarity and reproducibility, where the entire conversation history and the synthetic user's backstory are included within the prompt context for generating each subsequent utterance (\cref{appx:sleep_prompt}) \footnote{A code snippet for instantiating our synthetic users using Concordia is available at \url{https://anonymous.4open.science/r/sleepless_nights}}.

\begin{figure}[t]
  \centering
  \includegraphics[width=\columnwidth]{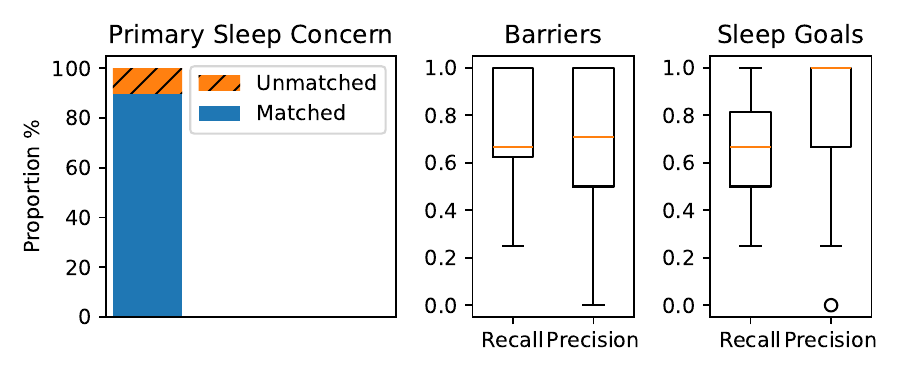}
  \caption{\textbf{Extracting synthetic users' sleep profiles from interaction for automated evaluation of the coaching agent}. Our synthetic users ($N$=68) with sleep conditions interacted with our sleep agent for 10 turns. Recall and precision were computed by comparing the internal state-of-user model in the coaching agent to the true sleep profile of the synthetic user. ``Primary sleep concern'' is a single natural language statement, while ``barriers'' and ``sleep goals'' include multiple natural language items.}
  \label{fig:reasoner_sleep_profile}
  \vspace{-1.4em}
\end{figure}

\subsection{The Sleep Coaching Agent}

We used a sleep coaching agent based on \citet{christakopoulou2024agentsthinkingfastslow}, which proposed a two-agent system consisting of a ``Talker'' and a ``Reasoner'' agent. Inspired by \citet{kahneman2011thinking}, the Reasoner (``System 2'') is responsible for generating an internal model of the use based on conversation history, in addition to planning and calling tools, while the Talker  (``System 1'') is responsible for generating conversation based on the structured state generated by the Reasoner. We used Gemini 1.5 Pro \citep{geminiteam2024gemini15unlockingmultimodal} as the LLM engine for both constituent agents.

\subsection{Evaluating Agent-Synthetic User Interactions in Sleep Coaching}
\label{subsec:sleep_eval}

Explicit modeling of the user in the sleep coaching agent enabled us to compare the internal user state to the ``true'' sleep profile we  explicitly assigned to the synthetic user. After a 10-turn interaction between a synthetic user and the coaching agent (the turn length of 10 was chosen to balance efficiency and comprehensiveness, after reviewing real sleep coach-user interactions in \citet{Perlis2005-jw}), the coaching agent was able to identify the synthetic user's primary sleep concern with 89.7\% accuracy (\cref{fig:reasoner_sleep_profile}; error analysis in \cref{appx:error_analysis}). For barriers and sleep goals, which are also explicitly modelled by the coach agent and consist of multiple items (natural language sentences) per user, we computed recall (sensitivity) and precision, as measured by the proportion of the true attributes also captured by the coaching agent's model, and the proportion of the attributes in the coaching agent's model that were present in the true attributes, respectively. When matching sentences, we used fuzzy-matching by a high-performance, instruction-tuned LLM (Gemini 1.5 Pro; \citet{geminiteam2024gemini15unlockingmultimodal}), to account for paraphrasing (e.g., ``inconsistent sleep duration'' and ``variable sleep duration''; \cref{appx:sleep_profile_fuzzy_matching}), and manually checked a random subset  (10 individuals) to make sure the answers were consistent with human answers (an easy task for modern LLMs; see, for example, \citet{Michail2025-sx}). We obtained 71.4\% mean recall and 72.5\% mean precision for the barriers, and 66.4\% mean recall and 84.2\% mean precision for the sleep goals (\cref{fig:reasoner_sleep_profile}). We also conducted the same experiment using Gemma 2 (``27B-IT'' version) \citep{Gemma-Team2024-rj} as an alternative open-source backend LLM for synthetic user generation in Concordia, and observed performance degradation (\cref{appx:sleep_profile_gemma}).

\begin{figure}[t]
  \centering
  \includegraphics[width=0.8\columnwidth]{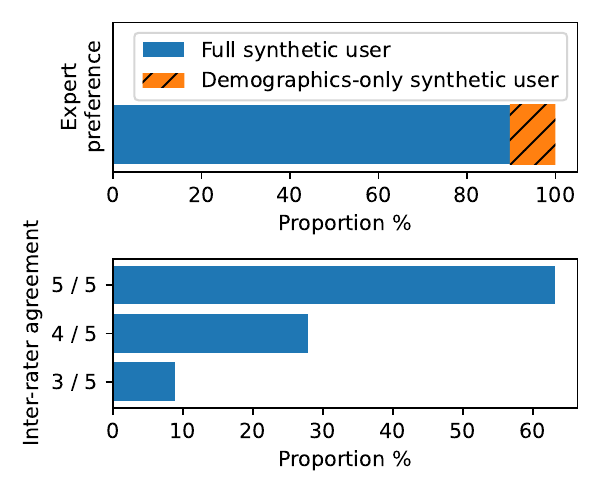}
  \caption{\textbf{Expert evaluation of grounded synthetic users}. Human expert evaluation of interactions from two sets of synthetic users, one from the full synthetic user pipeline and another from demographics-only data without sleep conditions or behavioral/psychological characteristics, with a fixed sleep coaching agent. Five human experts annotated each case independently.}
  \label{fig:sleep_human_eval}
  \vspace{-1.4em}
\end{figure}

Finally, to validate our use of synthetic users with health and behavioral/psychological attributes in addition to basic demographic information, we conducted a human expert evaluation of interactions from two synthetic users: one from our full synthetic user pipeline and another from using demographics-only data without health conditions, with the same sleep coaching agent. We asked human evaluators, with training and quality control conducted by a clinical psychologist, to annotate their preference between two interactions with each synthetic user, with 5$\times$ coverage of each question (\cref{appx:sleep_interaction_example}). The evaluators overwhelmingly favored our full synthetic users over the baseline and the differences were highly statistically significant ($p\textrm{-value} = 3.7\times 10^{-12}$ by one-tailed test under binomial null distribution), with high inter-rater reliability (64\% of the cases achieving the complete 5/5 agreement and 91\% of the cases achieving 4/5 agreement or more; Fleiss' $\kappa = 0.67$ \citep{Fleiss1971-iy}) (\cref{fig:sleep_human_eval}).

\section{Experiments: Diabetes Coaching}\label{sec:diabetesCoaching}

Diabetes mellitus, commonly referred to as diabetes, is a highly prevalent chronic disease, affecting 15\% of all US adults in 2021 \citep{CDC2024-io}. Understanding lifestyle barriers is crucial for effectively addressing the challenges this condition presents \citep{Deslippe2023-oy}. The barriers significantly impact an individual's ability to adhere to recommended behaviors and achieve optimal health outcomes. Many barriers are intricately driven by demographic, socioeconomic, and clinical circumstances, collectively manifesting into specific challenges. By accurately representing these interconnections, we can create synthetic user profiles that faithfully reflect the real-world complexities encountered by individuals with cardiometabolic conditions, thereby enhancing the realism and applicability of our synthetic users interacting with a diabetes coaching agent (\cref{fig:overview}).

\subsection{Synthetic Users with Diabetes}

To obtain a representative distribution of cardiometabolic barriers, we leveraged  insights from 100 peer-reviewed articles in \citet{yang2024barrierstacticsbehavioralscienceinformed}, which comprehensively identified 246 challenges encountered by cardiometabolic patients, in consultation with behavioral experts. The challenges were classified into six distinct sub-categories within the COM-B (capability, opportunity, motivation - behavior) model, a widely recognized framework for understanding behavior in healthcare \citep{Michie2011-gp}. The challenges were again categorized into 21 distinct barrier concepts by behavioral experts, each belonging to a sub-category within the COM-B model, as shown in \cref{appx:comb}. Through this process, we ensured that the synthetic user profiles developed for simulated coaching agent interactions were representative, reflecting the appropriate distribution of challenges.

To derive our synthetic users with diabetes, we utilized Project Baseline Health Study (PBHS), a longitudinal cohort with diverse backgrounds representative of the entire health spectrum \citep{Arges2020-zh}.\footnote{The data use was formally approved by our institution's internal review committee, ensuring ethical compliance.} After careful preprocessing (\cref{appx:pbhs_diabetes_preprocessing}), we obtained a cohort of 345 Type 2 diabetic individuals with diverse demographic, social, medical, and health attributes (\cref{appx:pbhs_diabetes_data}).

To construct realistic vignettes of synthetic users from this cohort, we first sampled a COM-B category under its original distribution in PBHS (\cref{appx:comb_dist}), and then uniformly sampled a specific barrier within the selected COM-B category. Using this identified barrier, we randomly selected a corresponding individual from PBHS with relevant symptoms. The selected individual's demographic, socioeconomic, clinical, and behavioral survey data formed the foundation of our natural language vignette (\cref{fig:overview}). To enhance the narrative quality, realism, and depth of these vignettes, a coherent backstory was generated by an LLM, incorporating the identified barrier (\cref{appx:diabetes_nutrition_vignette_prompt}). Finally, we generated a communication style field (e.g., tone, verbosity, and confidence) to compose the final vignette. Notably, the specific technical term for the selected barrier was not included in the vignette, since human users are unlikely to label themselves under it. In total, 200 vignettes for synthetic users were generated using the commercially available Gemini 2.0 Flash model chosen for its reasoning and conversation capabilities \citep{Pichai_2024}.

\subsection{The Diabetes Coaching Agent}
The synthetic users generated from the vignettes interacted with a diabetes coaching agent, building upon methodologies developed by \citet{yang2024barrierstacticsbehavioralscienceinformed}. This coaching agent was constructed to assist users to set health goals and overcome identified barriers. After each simulated interaction, the coaching agent was asked to identify one of the 21 barriers in its modeling of the user. Both the synthetic users and the coaching agent were generated using Gemini 1.5 Pro \citep{geminiteam2024gemini15unlockingmultimodal}.

\subsection{Evaluating Agent-Synthetic User Interactions in Diabetes Coaching}
\label{subsec:diabetes_eval}

\begin{figure}[t]
  \centering
  \includegraphics[width=0.9\columnwidth]{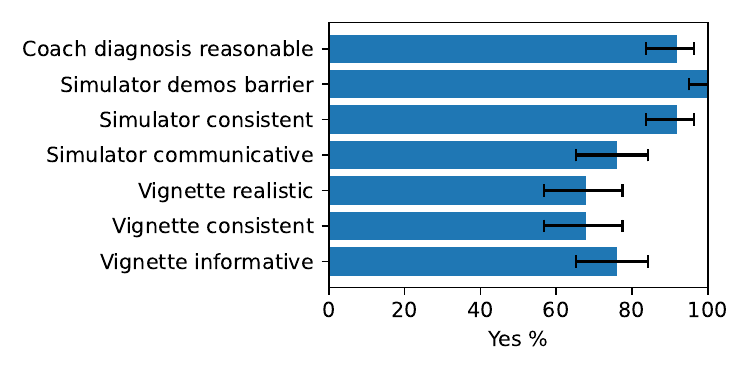}
  \caption{\textbf{Assessing synthetic users' performance from interaction with the diabetes coaching agent}. Three human annotators reviewed 25 randomly-selected simulated interactions. Wilson's confidence interval was used for these binary annotation tasks.}
  \label{fig:cardio_exp_labels}
  \vspace{-1.5em}
\end{figure}

To assess the fidelity and reliability of the synthetic users, we solicited evaluations from a panel of three experts, comprising behavioral scientists and patient care practitioners (\cref{appx:diabetes_eval}). Each expert received  25 randomly-selected simulated interactions (out of 200) for assessment. Experts agreed that the synthetic users were highly consistent (92\%) and effectively demonstrated the barriers they were designed to portray (100\%) (\cref{fig:cardio_exp_labels}; \cref{appx:diabetes_eval_results}).

Similarly to our sleep coaching evaluation experiments (\cref{subsec:sleep_eval}), we also conducted a secondary experiment comparing our synthetic users to baseline synthetic users generated from general demographic data only, without the PBHS data and the COM-B barriers that our methodology uniquely provides, in their interactions with the same coaching agent. After randomly selecting an individual from PBHS, the baseline synthetic user was generated from their standard demographic information, while our full synthetic user had the same standard demographics, richer health data, and the individual's original barrier (\cref{appx:diabetes_nutrition_patient_agent_prompt}). Both synthetic users were asked to portray an individual with cardiometabolic health challenges, and we ensured both possessed the same standard demographic data, communication style, verbosity, and confidence, for a fair comparison.

\begin{figure}[t]
  \centering
  \includegraphics[width=\columnwidth]{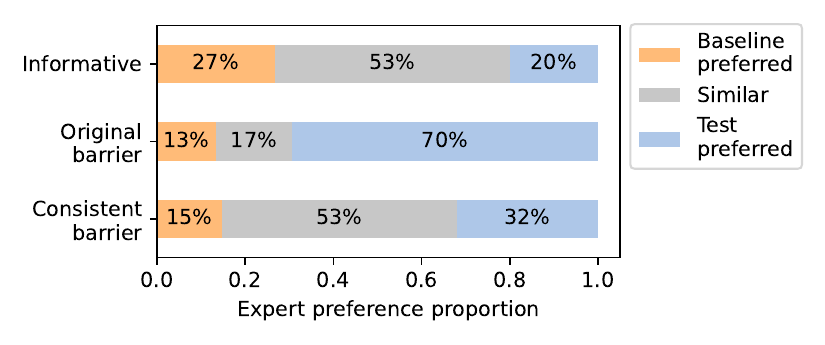}
  \caption{\textbf{Comparing grounded synthetic users to a baseline on simulated diabetes coaching conversations.} Human annotators reviewed 75 pairs of conversations, and labeled which conversation demonstrated the dimensions represented in each row. ``Test'' represents our grounded synthetic users and ``baseline'' represents synthetic users created from basic demographics only.}
  \label{fig:diabetes_comparison}
  \vspace{-1.5em}
\end{figure}

Human annotators were assigned three evaluation questions (\cref{appx:diabetes_eval_comparison_qs}) to annotate randomly selected interactions without actual vignettes, testing implicit demonstration of the vignettes through conversations. 75 pairs of interactions (full synthetic users and baseline synthetic users) were evaluated (\cref{fig:diabetes_comparison}).

For consistent representation of a single barrier (correct or not), annotators preferred the interactions from our full synthetic users twice as much as the baseline (32\%:15\%). Moreover, the annotators overwhelmingly preferred the full synthetic user (70\%) for demonstrating the \emph{correct} original barrier. We observed mixed results for informativeness, with the baseline being slightly preferred and more than half of the interaction pairs annotated as ``Similar''. Overall, this analysis demonstrated that our synthetic users can more accurately portray the original barrier consistently in simulated interactions (\cref{fig:diabetes_comparison}), which has an important downstream effects. This implies our synthetic user pipeline can be used to generate training data for efficiently fine-tuning the coaching agent for a specific barrier for incremental improvement. On the other hand, simulated interactions with the baseline synthetic users would not reliably represent the specific barrier one seeks to better capture.

Finally, we investigated the difference in distribution of barriers that the baseline synthetic users (using only demographic information) created, compared to the reference distribution observed in the literature (\cref{appx:barrier_butterfly}). As the baseline synthetic users only express barriers through conversation, we used the coaching agent's internal state (``diagnosis'') of the user's barrier established through interaction, which was evaluated as credible by human experts (\cref{fig:cardio_exp_labels}). Interestingly, the baseline synthetic users significantly over-sampled certain barriers (physical capability, reflective motivation), while under-sampling others (social opportunity, physical opportunity), highlighting the importance of grounding synthetic users in behavioral data from real human users whom the coaching agent is designed to serve.

\section{Conclusion}

We designed an end-to-end framework for generating synthetic users for evaluating coaching agents grounded in health, lifestyle, behavioral, and psychological attributes complementing basic demographics from human data. The additional sampling process was employed to ensure that the distribution of these attributes could be explicitly controlled (i.e., to ensure applicability to less represented sub-populations). Realistic backstories and additional health conditions (where not available in human user data) were conditionally generated from the grounded data. The final vignettes generated by this framework were used in a generative agent-based model framework to effectively simulate interactions between the synthetic users and the given coaching agent. We evaluated our methods in two independent, highly prevalent health coaching use cases of sleep coaching and diabetes management.

Efficient development of autonomous agents is significantly accelerated by evaluation metrics that can be computed \emph{without} complete dependence on human users, since human evaluation is costly and time-consuming.  While there is a large body of literature on the design of general-purpose synthetic users in this setting (see \cref{sec:background}), we have highlighted the importance of grounding synthetic users' health and behavioral attributes on a real human dataset, with concrete demonstrations and evaluations of our end-to-end pipeline in two independently developed coaching agents. Although some of our demonstrations relied on access to the coaching agents' internal user model, the simulated interactions generated by our framework can be evaluated in a way that is agnostic to the design of the agent, so the framework can be applied to any conversational agents.

The extensive, realistic, and grounded simulated interactions produced by our framework lay the foundation for efficient development of coaching agents, potentially through a fine-tuning or reinforcement learning loop, as will be pursued in future research.

\section*{Limitations}

This work introduced an end-to-end framework for evaluating a coaching agent using automated synthetic users. The ultimate goal of performing this evaluation is to improve the coaching agent itself using the metrics and signals derived from this work, which is left for future research. The coaching agents we evaluated in this work may not have the ``state-of-the-art'' performance, as that was not the focus of this paper. We also did not compare multiple commercial LLMs, as that was also not the focus of this paper.

Using LLM-powered agents presents risks for users. One of the motivations for this work is characterizing and minimizing some of those risks by enabling predictive agent evaluation with realistic user personas.

In evaluating coaching agents, a longitudinal assessment of potential behavioral change is crucial. Effectively and realistically simulating synthetic users' behavioral change is a critical open question to be addressed, including modeling non-adherence to coaching advice (e.g., human users may not follow the advice, or may \emph{claim} they changed their behavior, even if they did not).

Personalization and adaptation of coaching and other agents are very important and relevant topics that are beyond the scope of this paper. We hope that grounded, rich, and diverse personas of synthetic users we developed in this work will significantly support the process of evaluating and improving personalized agents adapting to evolving user behaviors and needs.

\clearpage
\bibliography{ref}

\clearpage
\onecolumn
\appendix

\renewcommand\thefigure{A\arabic{figure}}
\setcounter{figure}{0}

\renewcommand{\thetable}{A\arabic{table}}
\setcounter{table}{0}


\section{LifeSnaps Dataset Preprocessing}
\label{appx:lifesnaps_detail}
The age and the gender of the participants in the LifeSnaps dataset were available in two categories: ``under 30'' and ``over 30'', and ``male'' and ``female''. Of the 71 participants, we excluded 3 individuals whose age or gender data were missing in the dataset, resulting in 68 individuals. BMI was classified into four categories following a standard practice and terms: ``underweight'' ($\textrm{BMI} < 19$), ``normal'' ($19 \leq \textrm{BMI} < 25$), ``overweight'' ($25 \leq \textrm{BMI} < 30$), and ``obese'' ($\textrm{BMI} \geq 30$).

Given longitudinal LifeSnaps data over four months, we computed the mean and the standard deviation of sleep duration. The mean sleep duration was included in the vignettes in the number of hours. Participants with whose standard deviation of duration was no more than 1.5 hours were considered to have ``consistent'' sleep duration, and the rest were considered to have ``variable'' sleep duration.

\section{Synthetic User Prompt Generation for the Sleep Domain}
\label{appx:sleep_prompt}

Prompt components: \texttt{\cbBlue{Vignette}}, \texttt{Context}, and \texttt{\cbBurgundy{Instructions}}. A \texttt{Context} consists of the synthetic user's \texttt{\cbGreenSea{Name}} and the entire \texttt{\cbRose{conversational transcript}} so far.

\begin{quote}
\texttt{\cbBlue{Nicole is female. She is more than 30 years old. She typically sleeps for 8 hours at night. Her sleep duration is highly variable. Her sleep efficiency is typically at 94\%. She is underweight. She is neither introverted nor extraverted. She is highly agreeable. She is moderately conscientious. She feels unstable. She is highly intellectual. (...)} \\
You are \cbGreenSea{Nicole}. \\
Your last observations are: \\
\cbRose{Nicole -- ``I'm struggling with getting enough sleep.  I'd like to sleep 7-8 hours a night consistently so I can improve my energy levels at work and be more present for my family, but my demanding work schedule and racing thoughts make it difficult.'' \\
COACH -- ``It's completely understandable that a demanding work schedule and racing thoughts can make sleep a challenge. Wanting better energy and to be more present for your family are fantastic goals! Is there something specific you've tried in the past to help with sleep, or something you'd like to try now?'' (...)} \\
\cbBurgundy{Given the above, generate what Nicole would say to the COACH next in this conversation. Respond in the format ``Nicole -- ...''. If the COACH has asked for input, make sure to generate specific and detailed answer to the question. Bear in mind the primary sleep concern, sleep goals, reasons for those goals, and barriers of Nicole provided above. The tone and style of the conversation should math Nicole's descriptions above.}
}
\end{quote}

\section{Sleep Profile Generation}
\label{appx:sleep_profile}

The following prompt was used to generate sleep profiles for synthetic users based on LifeSnaps, where [name] is randomly sampled and [background\_info] is generated by real data.

\begin{quote}
Here is some background information about a fictional person, called [name]. I want to create a fictional profile of [name]. The sleep profile should be in a structured JSON format, including these fields: "primary\_sleep\_concern", "sleep\_goals", "reasons\_for\_goals", "barriers". "primary\_sleep\_concern" must be a single string. "sleep\_goals" must be a list of strings. "reasons\_for\_goals" must be a list of strings. "barriers" must be a list of strings. Be creative and make up stories if you need to, but make sure the JSON sleep profile is consistent with the given background information. Your response must start with a justification of each field you create, followed by the structured JSON. It's important that you only use escaped double quotes and single quotes, as well as escaped backslashes to not break the JSON format. It's also important that you respect the JSON format.

Background information: "[background\_info]"

Response:
\end{quote}

\section{Prompts for Fuzzy Matching Sleep Profiles}
\label{appx:sleep_profile_fuzzy_matching}

For fuzzy-matching two descriptions of primary sleep concerns to account for paraphrasing, the following prompt was used for an instruction-tuned high-performance LLM.

\begin{quote}
I would like you to answer this question.

The following are two descriptions of someone's primary sleep concern, A and B:

A: "[description A]"

B: "[description B]"

Does A and B generally communicate compatible sleep concerns? They do not have to match completely, but I want to know whether they GENERALLY describes similar concerns. It is okay if some specific details are different.

Start your answer with Yes or No, followed by an explanation.
\end{quote}

For fuzzy-matching a list of sleep attributes (such as sleep goals and barriers), the following prompt was used to generate precision and recall.

\begin{quote}
I would like you to answer two questions. I have two lists of [field name] of a person. I want to know how they match well with each other. There are [length of list A] items in the list A, and there are [length of list B] items in the list B:

A: [list A]

B: [list B]

My two questions are: How many of the items in the list A are also generally represented in the list B? And how many items in the list B are also generally represented in the list A?

When finding a match, not all the details need to match. I just want to know whether each item is generally represented in the other list, while some specific details may be different.

Start your answer with two numbers separated by a comma and followed by a period, representing the answers to the two questions, for example "3, 2.". Add your explanation after this answer.
\end{quote}

\section{COM-B Categories}
\label{appx:comb}

\begin{center}
\begin{tabular}{p{0.25\columnwidth}|p{0.75\columnwidth}}
    \hline
    \textbf{COM-B Category} & \textbf{Barriers} \\
    \hline
    Psychological Capability & Don’t know the Basics, Don’t know the consequences, Planning fallacy, Memory \\ 
    \hline
    Physical Capability & Decision fatigue, Physical limitations \\  
    \hline
    Social Opportunity & Lack of social support, Conflicting opinions, Impact on others, peer pressure \\
    \hline
    Physical Opportunity & Geographic limitations, Affordability or costs, Lack of equipment, Switching settings \\
    \hline
    Reflective Motivation & Poor self-efficacy, Competing priorities , Lack of desire without reasons, Boredom \\
    \hline
    Automatic Motivation & Present bias, Anchoring effect, Gut feelings \\
    \hline
\end{tabular}
\end{center}

\section{COM-B Category Distribution}
\label{appx:comb_dist}

The COM-B sub-category distribution was 25\% reflective motivation, 21\% psychological capability, 19\% physical opportunity, 15\% social opportunity, 12\% automatic motivation, 9\% physical capability. This relative distribution of the barriers were also observed in related studies \citep{MacPherson2023-ox}.

\section{PBHS Diabetes Data Preprocessing}
\label{appx:pbhs_diabetes_preprocessing}

First, we filtered the PBHS dataset to include only individuals diagnosed with type 2 diabetes, resulting in a cohort of 345 individuals. For each individual, we leveraged their comprehensive demographic, socioeconomic, clinical, and behavioral survey data. The data encompassed a range of variables, including age, sex, race, marital status, smoking status, education level, income, insurance coverage, household size and structure, social interactions (measured through gatherings, phone calls, and text messages), attendance at organizational meetings, and employment status. Additionally, clinical parameters such as diagnosed conditions, Hemoglobin A1c (HbA1c) levels, blood glucose measurements, (systolic/diastolic) blood pressure, and body mass index (BMI) were used. Furthermore, the survey data underwent manual coding to identify if any of the individuals reported experiencing any relevant symptoms of the 21 identified barriers as their primary challenges. This coding process ensured that the nuanced, self-reported experiences of the individuals were captured and integrated into the development of our simulated user profiles.

\section{PBHS Diabetes Data Attributes}
\label{appx:pbhs_diabetes_data}
The following fields from PBHS data were used as input for synthetic user vignette generation (see \cref{appx:diabetes_nutrition_vignette_prompt}. 

\begin{center}
\begin{tabular}{p{0.3\columnwidth}|p{0.7\columnwidth}}
    \hline
    \textbf{Data Category} & \textbf{Attribute} \\
    \hline
    Basic demographics & Age at enrollment \\ 
    \hline
    Basic demographics & Race \\  
    \hline
    Basic demographics & Marital status \\
    \hline
    Basic demographics & Education \\
    \hline
    Basic demographics & Income \\
    \hline
    Basic demographics & Employment status \\
    \hline
    Social environment & People living at home \\
    \hline
    Social environment & Number of people under 18 living at home \\
    \hline
    Social environment & Weekly number of friend and family gatherings \\
    \hline
    Social environment & Weekly number of phone calls with friends and family \\
    \hline
    Social environment & Weekly number of texts with friends and family \\
    \hline
    Social environment & Weekly attendance to social organization meetings \\
    \hline
    Health data & Smoking status \\ 
    \hline
    Health data & Has insurance \\
    \hline
    Health data & Diagnostic conditions \\ 
    \hline
    Medical measurements & Hemoglobin A1C (HbA1c) \\
    \hline
    Medical measurements & Blood Glucose (mg/dl) \\
    \hline
    Medical measurements & Systolic and diastolic blood pressure \\
    \hline
    Medical measurements & Body Mass Index (BMI) \\
    \hline
    Barrier & Matched barrier from PBHS data \\
    \hline
\end{tabular}
\end{center}

\section{Synthetic User Prompt for the Diabetes Domain}
\label{appx:diabetes_nutrition_patient_agent_prompt}

Prompt components: \texttt{\cbRose{Data}}, \texttt{\cbBlue{Vignette}}, and \texttt{\cbBurgundy{Instructions}}. \texttt{\cbRose{Data}} is the synthetic user's demographic, social, and medical data from PBHS. \texttt{\cbBlue{Vignette}} is the backstory of synthetic user's barriers and \texttt{\cbRose{Data}} summarized in a few sentences. Conversation history is tracked and accounted for separately through a stateful chat session to ensure synthetic user is consistent.

\begin{quote}
\texttt{\cbBurgundy{You are a patient with cardiometabolic condition using a digital chat based application, below is your specific role. Pursue the conversation and the AI Coach will work with you. Reference your patient details and bring up information that can be relevant to the conversation. ONLY write out the conversation (not the breathing nor internal thoughts).\\} \\
Patient details: \cbRose{[PBHS data (..)]\\} \\
\cbBlue{John is a 31-year-old working dad, with a busy life. He lives with his partner and three young children and is trying to juggle work and family life. He knows he needs to eat better to manage his diabetes with an HbA1c of 6.7 and a high blood glucose level of 305, and he genuinely wants to improve. He often tells himself he will eat a healthy lunch, but finds himself eating fast food because he didn't have time to pack a lunch or get the right groceries the night before. He knows he's not doing his best, and feels like he is letting himself down, but also feels overwhelmed and unsure of where to even begin.\\} \\
\cbBurgundy{Now it's your turn to speak, you are the patient. You should follow your role and continue the conversation based on the AI Coach's last message. \\
- Make sure you focus on the backstory section of the patient details - specifically the BARRIER the patient is facing.\\
- Use the other patient details as supporting information in a natural, smooth way if necessary. When using those details, make sure to talk like a regular patient (change medical jargon to common speak) and not a doctor. \\
- DO NOT hallucinate or make up any new information that are not in the patient details. Stick to the context in the backstory presenting the barrier / challenge. \\
- Make sure to follow the communication style given in the patient details. \\ 
- You should only respond with at most 2 sentences per turn. \\
}
}
\end{quote}

\section{Diabetes Synthetic User Vignette Generation}
\label{appx:diabetes_nutrition_vignette_prompt}

\subsection{Initial Vignette Generation}
Prompt components: \texttt{\cbRose{Data}}, \texttt{\cbLilac{Barrier}}, \texttt{\cbGreenSea{Few Shot Examples}}, and \texttt{\cbBurgundy{Instructions}}. \texttt{\cbRose{Data}} is the synthetic user's demographic, social, and medical data from PBHS. \texttt{\cbGreenSea{Few Shot Examples}} are example synthetic user vignettes in JSON format to guide vignette generation. \texttt{\cbLilac{Barrier}} is the barrier the synthetic user is facing randomly selected from a prior distribution. The prompt below is for our test vignette; for the baseline vignette, we exclude health data and attributes unique to PBHS.

\begin{quote}
\texttt{\cbBurgundy{You are an expert in creating realistic patient vignettes for a digital health app with an AI coach providing goals achievement support for patients with diabetes. You are tasked with creating a realistic vignette for a patient working on their health goals. Given a barrier that a patient is experiencing, fill in the the following possible parameters that are aligned to each other and to the barrier, realistic for a diabetes patient, and collectively and informatively represent the barrier.\\}
\\
Here is the barrier the patient is experiencing: \texttt{\cbLilac{[Barrier]}}\\
\\ 
Here are the possible parameters you can fill in: \texttt{\cbRose{[PBHS Data (...)]}}  \\ 
\\ 
\texttt{\cbBurgundy{Include the following additional parameters as well:\\ 
- Name: Patient's name matching the patient SEX provided.\\ 
- Tone: (formal, academic, casual, playful, agreeable, antagnoistic, resistant, depressed, apathetic)\\ 
- Verbosity: (Shares intentional, complete sentences, responds to each question asked; Responds in short sentences or phrases; Shares unrelated information / overshares)\\ 
- Confidence: (High confidence and self awareness, knows themselves and conveys accurately; Concerned for appearance, erring towards aspirational self, overly optimistic view of oneself; Low confidence, convinced they are likely doing something wrong, apologetic)\\ 
\\ 
Here are the rules for creating the vignette: \\ 
- First, fill in each field of the parameters with a realistic value that is aligned with the barrier. The values can be numerical values or short free text descriptions. \\ 
- Then, provide a short backstory paragraph that describes the patient's situation, including the filled in parameters. The paragraph should be realistic and informative to represent the barrier. \\ 
- IMPORTANT: It is critical that the specific barrier term is not used in the vignette, including the backstory. The vignette should represent the barrier without explicitly mentioning it. \\ 
\\ }}
Here is an example of a vignette with the barrier of \texttt{\cbLilac{[Barrier]}}: \\
\texttt{\cbGreenSea{[Few shot examples]}}\\ 
\\ 
\\ 
\texttt{\cbBurgundy{Your final output should be the following, in JSON format.\\ 
\{\{\\ 
    ``reasoning'': <Your reasoning behind the vignette you are generating>,\\ 
    ``vignette'': <The vignette you have created in valid python dictionary format with keys containing the relevant parameter fields with an additional key containing the overall backstory.>\\ 
\}\}\\ 
Think step by step, and validate your reasoning with your text.
}}}
\end{quote}

\subsection{Vignette Generation Improvement}
After initial generation, we developed another ``verifier LLM'' to improve the quality of the vignettes. This verifier LLM edited components of the vignette and backstory to ensure that they were realistic and consistent with each other. This idea of ``self-refinement'' was inspired by existing literature \citep{madaan2023selfrefineiterativerefinementselffeedback,yao2023reactsynergizingreasoningacting}, but we used a separate LLM with fewer iterative steps. The prompt used by our verifier LLM is shown below.

Prompt components: \texttt{\cbBlue{Vignette}}, \texttt{\cbLilac{Barrier}}, \texttt{\cbGreenSea{Few Shot Examples}}, and \texttt{\cbBurgundy{Instructions}}. \texttt{\cbBlue{Vignette}} is the initially generated synthetic user vignette to improve. \texttt{\cbGreenSea{Few Shot Examples}} are example synthetic user vignettes in JSON format to guide vignette generation. \texttt{\cbLilac{Barrier}} is the barrier the synthetic user is facing randomly selected from a prior distribution. 

\begin{quote}
\texttt{\cbBurgundy{You are an expert in realistic patient vignettes for a digital health app with an AI coach providing goals achievement support for patients with diabetes. You are tasked with improving / editing an existing realistic patient vignette provided to you.\\ \\
Use the following methods to improve the vignette and parameters:\\
1) Remove information that causes the vignette to be unrealistic or misaligned with the barrier.\\
2) Ensure the vignette is representative to ONLY the barrier provided to you, without explicitly mentioning the barrier.\\
3) Add information that makes the vignette more realistic.\\
4) Double check that the backstory is using the correct age field and not hallucinating.\\
5) Remove any unnecessary formatting (curly brackets, hallucinations, etc).\\} \\
Here is the barrier the patient is experiencing: \cbLilac{[Barrier]} \\ \\
\cbBurgundy{Your final output should be the following, in JSON format.\\
\{\{\\
    "reasoning": <Your reasoning behind how you improved the vignette>,\\
    "vignette": <The vignette you have improved in valid python dictionary format with the exact same keys as the provided vignette.>\\
\}\}\\
Think step by step, and validate your reasoning with your text.\\
}\\
Improve the following vignette: \cbBlue{[Vignette]}\\ \\
EXAMPLES:\\
\cbGreenSea{[Few Shot Examples (...)]}\\
}
\end{quote}

\section{Diabetes Evaluation}
\label{appx:diabetes_eval}

Human experts were asked to provide insights by responding to the following evaluative questions:

\begin{enumerate}
    \item The vignette informatively represents the specified barrier. (Y/N) 
    \item The information in the vignette is self-consistent. (Y/N)
    \item The traits in the vignette are realistic for a cardiometabolic patient. (Y/N)
    \item The patient simulator's communication style specified in the vignette is reflected in the conversation. (Y/N)
    \item The patient simulator’s responses are consistent with the vignette. (Y/N)
    \item The patient simulator informatively demonstrates the barrier specified in the vignette. (Y/N)
    \item The coaching agent’s diagnosis of patient barriers is reasonable given the conversation alone. (Y/N)
\end{enumerate}

\section{Diabetes Evaluation Results}
\label{appx:diabetes_eval_results}

\begin{center}
\begin{small}
\begin{tabular}{p{5.5cm}|p{1.75cm}|p{1.75cm}|p{1.75cm}|p{1.75cm}}
\hline
\textbf{Question} & \textbf{Expert 1 Yes Response Rate} & \textbf{Expert 2 Yes Response Rate} & \textbf{Expert 3 Yes Response Rate} & \textbf{All Experts Yes Rate} \\
\hline
Q1: Vignette is informative & 0.88 & 0.92 & 0.84 & 0.76 \\
\hline
Q2: Vignette is self consistent & 0.72 & 0.88 & 0.80 & 0.68 \\
\hline
Q3: Vignette is realistic & 0.88 & 0.76 & 0.96 & 0.68 \\
\hline
Q4: Simulator portrays correct communication style & 0.88 & 0.88 & 1.00 & 0.76 \\
\hline
Q5: Simulator is consistent with vignette & 0.96 & 0.96 & 1.00 & 0.92 \\
\hline
Q6: Simulator demonstrates barrier & 1.00 & 1.00 & 1.00 & 1.00 \\
\hline
Q7: Coach diagnoses reasonable barrier & 0.96 & 1.00 & 0.92 & 0.92 \\
\hline
\end{tabular}
\end{small}
\end{center}

\section{Diabetes Baseline vs. Test Comparison Questions}
\label{appx:diabetes_eval_comparison_qs}
Human experts were asked the following evaluation questions when comparing baseline vs. test set conversations. As previously described, the baseline is conversations generated using synthetic users from demographics-only data, while test synthetic users had the same demographics data, plus the rich health data and the COM-B barrier.

\begin{enumerate}[topsep=0pt,itemsep=-1ex,partopsep=1ex,parsep=1ex]
    \item Which of the two portrays one consistent barrier from the patient? (Y / N / Similar)
    \item Which of the two conveys the original barrier more accurately? (Y / N / Similar)
    \item Which of the two is more informative in understanding the patient? (Y / N / Similar)
\end{enumerate}

\section{Reference Distribution vs. LLM-Generated Distribution of Barriers in Simulated Conversations}
\label{appx:barrier_butterfly}
\begin{figure}[h]
  \centering
  \includegraphics[width=\columnwidth]{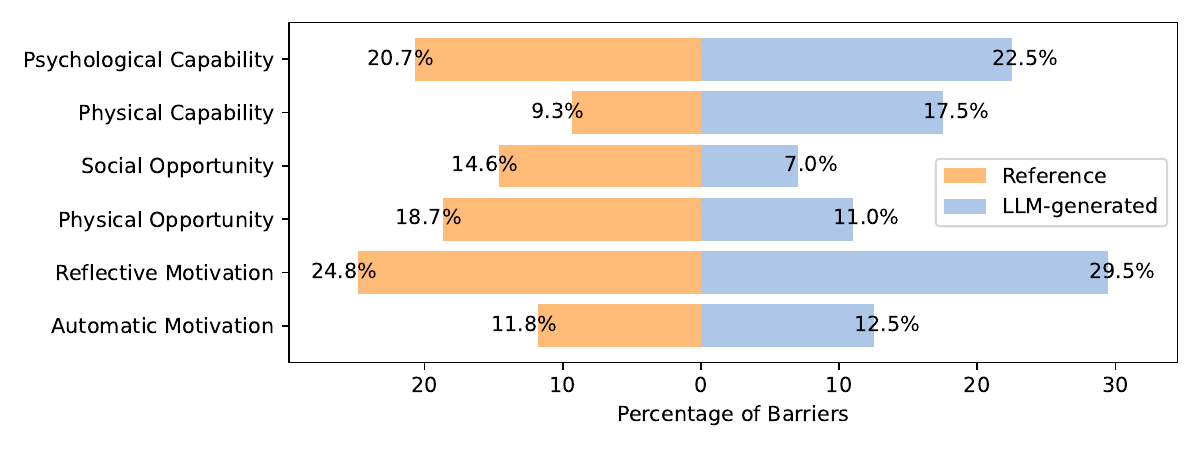}
  \caption{\textbf{Reference distribution vs LLM-generated distribution of barriers observed in simulated conversations}. The reference is the distribution of barriers observed in cardiometabolic behavioral literature. The LLM-generated distribution are extracted by the diabetes coaching agent in 200 simulated conversations.}
  \label{fig:barrier_butterfly}
\end{figure}

\section{Comparison With Gemma 2 IT 27B Model}
\label{appx:sleep_profile_gemma}
\begin{figure}[h]
  \centering
  \includegraphics[width=0.7\columnwidth]{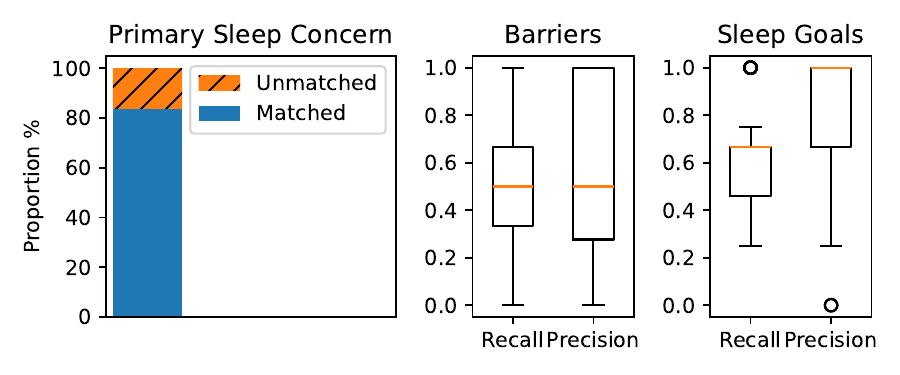}
  \caption{\textbf{Extracting synthetic users' sleep profile from interaction for automated evaluation of coaching agent, using Gemma 2 IT 27B instead of Gemini 1.5 Pro for synthetic user}. Our synthetic users ($N$=68) with sleep conditions interacted with our sleep agent for 10 turns. Recall and precision were computed by comparing the internal state-of-user model in the coaching agent to the true sleep profile of the synthetic user. ``Primary sleep concern'' is a single natural language statement, while ``barriers'' and ``sleep goals'' include multiple items, all in natural language. cf. \cref{fig:reasoner_sleep_profile}.}
  \label{fig:reasoner_sleep_profile_gemma}
\end{figure}

The following table includes a comparison of the key auto-evaluation metrics between synthetic users based on Gemma 2 IT 27B and Gemini 1.5 Pro.
\bigskip

\begin{tabular}{l|l|l}
\hline
\textbf{Metric} & \textbf{Gemma 2 IT 27B} & \textbf{Gemini 1.5 Pro}\\
\hline
Primary sleep concern accuracy \% & 83.8 & 89.7 \\
\hline
Barriers recall (mean) \% & 45.2 & 71.4 \\
\hline
Barriers precision (mean) \% & 55.5 & 72.5 \\
\hline
Sleep goals recall (mean) \% & 61.3 & 66.4 \\
\hline
Sleep goals precision (mean) \% & 82.6 & 84.2 \\
\hline
\end{tabular}

\section{Error Analysis and Failure Modes}
\label{appx:error_analysis}

We manually inspected the 10.3\% (7 out of 68) of the synthetic users from LifeSnaps for which our sleep coaching agent incorrectly determined their primary sleep concern. Qualitatively those 7 examples could be divided into two categories. One category was when the true primary concern and the coaching agent's prediction of the primary concern were related, but they did not pass the bar for paraphrase detection, such as ``maintaining consistent high-quality sleep'' vs. ``optimizing already good sleep''. The other category was when the synthetic users communicated multiple other concerns in addition to their primary sleep concern, and the coaching agent incorrectly determined one of the additional concerns as primary concerns, e.g. ``weight management and its impact on sleep quality'' vs. ``difficulty falling asleep due to stress, leading to low energy and impacting social life.'' We did not observe a complete failure of synthetic users in communicating their given issues.

\section{Agent Architecture Details}
\label{appx:agent_architecture_details}

Our sleep coaching agent followed the agent architecture in \citet{christakopoulou2024agentsthinkingfastslow}, which utilizes a novel Talker-Reasoner architecture and an explicit, structured model of the user. For our diabetes coaching agent, we used the agent developed by \citet{yang2024barrierstacticsbehavioralscienceinformed}, that probes for and identifies the root cause of an individual's dietary struggles (barriers) and delivers tailored tactics to overcome the identified barriers.

\section{Synthetic User-Sleep Coach Interaction Example}
\label{appx:sleep_interaction_example}

The following is an example interactions between the sleep coaching agent and two synthetic users we studied, the ``full'' synthetic user and a demographics-only synthetic user (\cref{subsec:sleep_eval}). It is also an example of the expert annotation questions we used to generate \cref{fig:sleep_human_eval}. For this particular question, the ``full'' synthetic user was option A, while the demographics-only synthetic user was option B.

\begin{quote}
\begin{small}

Question: The following is a description of a hypothetical person, who is interacting with a sleep coaching agent. Based on this profile, which one of interaction A and interaction B more closely matches the interaction between this person (=USER) and the agent (=COACH)?

Description: "William is male. He is less than 30 years old. He typically sleeps for 9 hours at night. His sleep duration is relatively consistent. His sleep efficiency is typically at 93\%. He is underweight. He is introverted. He is not agreeable. He is highly conscientious. He feels highly stable. He is moderately intellectual."

Note: What the COACH says is shown here to provide conversational context, but please ONLY use what USER says to determine your answer.

Answer: A / B ?

\bigskip

===== Interaction A =====

[USER]
I'm concerned about maintaining consistent sleep quality and duration, especially with increasing work demands.  I'd like to explore strategies for maintaining a 9-hour sleep schedule and optimizing my sleep environment.

[COACH]
Hi there!  It's great that you're prioritizing your sleep, especially with a busy work life. Maintaining consistent sleep can be a real game-changer!  So, what feels like the *biggest* hurdle for you in getting those 9 hours of quality sleep?

[USER]
It's a combination of things.  I find it hard to say no to extra work, which often leads to late nights.  Also, I'm sensitive to noise and temperature changes, so creating the ideal sleep environment is a challenge.

[COACH]
So, juggling work and creating a peaceful sleep sanctuary are your main concerns.  It sounds like saying "yes" to extra work sometimes means saying "no" to sleep. Can you tell me a bit more about your typical bedtime routine?

[USER]
I usually aim for a 10:30 PM bedtime to get my 9 hours, but it often ends up being closer to 11:00 PM or even later due to work.  I try to keep my room cool and dark, but I live in a noisy apartment building, which can be disruptive.

[COACH]
That 10:30 PM bedtime goal is excellent! It shows you're really committed to those 9 hours.  It sounds though, like work often pushes that back.  And the noisy apartment… yikes! Do you have any pre-sleep rituals that help you wind down, even when work runs late?

[USER]
I don't have a consistent wind-down routine, which might be part of the problem.  Sometimes I read before bed, but often I just collapse into bed exhausted after work.

[COACH]
So, sometimes reading, sometimes collapsing...it sounds like having a more consistent wind-down routine might be something to explore.  How would a good night's sleep improve your work life, do you think?

[USER]
A good night's sleep would definitely improve my focus and productivity at work, allowing me to handle those demanding tasks more efficiently.  It would also improve my overall mood and energy levels, making me more pleasant to be around, even though I'm naturally introverted.

[COACH]
More focus, better productivity, improved mood... those are *fantastic* benefits!  And hey, being more pleasant to be around is a win-win for everyone, introvert or not! If we could wave a magic wand and give you the *perfect* sleep, what would that look like for you?

[USER]
The perfect sleep would be consistently getting those uninterrupted 9 hours in a quiet, cool, and dark room, regardless of my workload.  Waking up feeling refreshed and energized every morning would be ideal.

[COACH]
Consistently getting 9 uninterrupted hours in a cool, dark, and quiet room, waking up refreshed... I love it!  That's a great vision! So, if we were to focus on just *one* thing to move you closer to that vision this week, what would it be?

[USER]
I think focusing on establishing a consistent wind-down routine would be the most impactful first step.  Perhaps something simple to start, like a short meditation or reading for a set amount of time before bed, even if work runs late.

[COACH]
That's a fantastic starting point!  A consistent wind-down routine can make a world of difference.  Short meditations or reading are both great options.  Even 10-15 minutes can signal to your body that it's time to sleep.  How about we aim for 15 minutes of either meditation or reading before bed every night this week?  We can also talk about strategies for dealing with the noise in your apartment, and tweaking your sleep environment to make it even more conducive to rest.  How does that sound?

[USER]
That sounds like a manageable commitment.  A consistent 15-minute wind-down routine before bed, even if it's just reading, seems like a good first step towards improving my sleep.

[COACH]
Great! I'm glad that feels doable.  Reading for 15 minutes is a wonderful way to signal to your body that it's time to rest.  To keep a consistent body clock, try reserving the hour before bed for activities that help you unwind. Simple, hands-on tasks like tidying up, doing dishes, or preparing your room for sleep are an easy and effective way to wind down after a long day. Repeating these simple tasks each night builds a routine that signals to your body when it's time to sleep.  We'll check in next week and see how it goes, and then we can talk about some strategies for creating a quieter sleep space.

[USER]
That sounds good.  I'll try the 15-minute reading routine before bed and reserving the hour before for winding down activities.  I appreciate the suggestions.

[COACH]
You've got this!  Remember, small, consistent steps lead to big changes.  I'm here to support you every step of the way.  Have a great week, and I look forward to hearing how it goes!

[USER]
Thanks. I'll let you know how the reading goes and if I have any trouble quieting down my apartment.

[COACH]
Sounds good! Remember, consistency is key. Even if some nights don't go perfectly, just keep coming back to that 15-minute wind-down routine.  I'm looking forward to hearing about your progress.

\bigskip

===== Interaction B =====

[USER]
I'm struggling with poor sleep quality, it's affecting my work.  I'd like to fall asleep faster, reduce nighttime awakenings, and wake up feeling refreshed so I can perform better at work, improve my mood, and have more energy for my hobbies, but I'm stressed about my work/life balance, have an inconsistent sleep schedule, and use electronics before bed.

[COACH]
It sounds like you're juggling a lot!  It’s totally understandable that you’re struggling with sleep when work, mood, hobbies, and everything in-between is on your mind.  So, out of all of those things, what feels like the *biggest* sleep challenge right now?

[USER]
I think the biggest challenge right now is my inconsistent sleep schedule because of social events and commitments.  It makes it really hard to stick to a regular sleep routine, which I know is important for good sleep.

[COACH]
It makes total sense that a wonky sleep schedule makes it tough to get good rest. Can you tell me a bit more about what a typical week’s sleep schedule looks like for you?  What time are you hitting the hay and waking up, even if it varies?

[USER]
Sure.  My bedtime and wake-up times vary quite a bit throughout the week. Weekdays, I usually try to be in bed by 11 PM, but with work events and social gatherings, it often ends up being closer to midnight or even later.  Then I have to wake up around 7 AM for work, so I rarely get a full night's sleep.

[COACH]
So if I'm hearing you right, weekdays are a bit of a sleep rollercoaster thanks to work and social stuff - sometimes bed by 11 p.m., sometimes midnight or later, but *always* up at 7 a.m.  What about weekends?  Is it any easier to stick to a schedule then?

[USER]
Weekends are a different story. I usually sleep in until 9 or 10 AM, sometimes even later if I've been out late the night before.  Then I stay up late, often past midnight, so my sleep schedule gets completely flipped.

[COACH]
So on weekends, you’re catching up on sleep by sleeping in, but then staying up late, which totally throws off your weekday rhythm.  It sounds like you’re dealing with “social jet lag,” where your sleep schedule does a 180 on the weekends. How do you think this shift affects you during the week?

[USER]
It makes it really hard to get back into a regular routine on Sunday night, so I'm often tired and sluggish on Mondays, which impacts my work performance.

[COACH]
Mondays sound like a rough start to the week thanks to that weekend sleep shift impacting your work.  If you *could* wave a magic wand and get your sleep sorted, what would the *perfect* sleep scenario look like for you?

[USER]
Ideally, I'd like to fall asleep around 10:30 PM every night and wake up feeling refreshed at 7 AM, even on weekends.  That way, I'd get a consistent 8-8.5 hours of sleep, which I think would make a huge difference in my energy levels and overall well-being.

[COACH]
So, the dream is 10:30 p.m. to 7 a.m. - consistent, even on weekends - giving you that solid 8 to 8.5 hours of ZZZs. I love it!  Do you have any ideas about how you might start moving towards that ideal bedtime and wake time, even if it seems difficult right now?

[USER]
I think I could start by setting a consistent bedtime alarm, even on weekends, to remind myself to start winding down.  Maybe even try shifting my bedtime 15 minutes earlier each night until I reach that 10:30 PM goal.

[COACH]
That’s a great idea!  Setting a consistent bedtime alarm, even on weekends, can really help regulate your body’s natural sleep-wake cycle.  Shifting your bedtime 15 minutes earlier each night is a smart approach, too – it’s gradual, so your body can adjust more easily.

While setting an alarm for *bedtime* is a fantastic first step, expert clinicians often suggest a slightly different approach to figuring out your *ideal* wake-up time. They recommend trying to wake up *without* an alarm for a week or two, even on weekends if possible, and noting down when your body naturally wakes up. This helps your body catch up on any sleep debt you might have, and it’ll give you a really good sense of how much sleep you truly need each night.  This natural wake-up time becomes your anchor point, and from there, you can count back 8 to 8.5 hours to find your ideal bedtime.

Once you have a better understanding of your natural sleep needs, you can then fine-tune your bedtime routine and create a relaxing wind-down ritual to help signal to your body that it’s time to sleep. How does that sound?  We can brainstorm some wind-down ideas together, too, if you’d like.

[USER]
That sounds like a good approach. I'm willing to try waking up without an alarm to see what my natural wake-up time is, and then work backward from there to figure out my ideal bedtime. I'd also appreciate some brainstorming on wind-down routines, as that's something I've struggled with.

[COACH]
Great! I’m happy to hear you’re open to trying this.  Let's talk wind-down routines.  A good wind-down routine is like telling your body, "Okay, it's time to shift gears and get ready for sleep."  Think of it as creating a relaxing and predictable pre-sleep ritual that you can look forward to.  Here are a few ideas we can explore, and we can tailor them to what you enjoy and what works for *you*:

*   **Dim the lights:**  Lowering the light levels in your home about an hour or two before bed helps your body start producing melatonin, the sleep hormone. Think soft lamps, candles (safely!), or even dimmer switches.
*   **Screen-free time:**  The blue light emitted from electronic devices like phones, tablets, and computers can interfere with melatonin production and make it harder to fall asleep.  Aim to put your screens away at least an hour before bed.
*   **Warm bath or shower:** A warm bath or shower can be incredibly relaxing. The rise and subsequent fall in body temperature after getting out of the water can actually help promote sleepiness.
*   **Relaxing activities:** Think about activities that help calm your mind and body.  Reading a physical book (not an e-reader!), listening to calming music, gentle stretching, or journaling are great options.
*   **Mindfulness and meditation:** Even a few minutes of mindfulness or meditation can help quiet a racing mind and reduce stress.  There are plenty of guided meditation apps available if you're new to this.
*   **Consistent sleep environment:** Make sure your bedroom is dark, quiet, and cool. A comfortable temperature, a good mattress, and cozy bedding can make a world of difference.

What sounds most appealing to you from this list?  Or are there other things you’ve tried in the past, or are curious about trying now, that we should add to the list?

[USER]
I like the idea of a warm bath or shower before bed, and I definitely need to work on the screen-free time.  I'm also curious about trying some mindfulness exercises, maybe a short guided meditation to help me relax.

[COACH]
Those are all fantastic choices! A warm bath or shower, screen-free time, and mindfulness exercises can be powerful tools for a relaxing wind-down routine.  

Let's break this down into smaller, manageable steps:

1.  **Screen-Free Time:**  Since you mentioned this is a big one, how about starting with just 30 minutes of screen-free time before bed?  You can gradually increase this time as you get more comfortable. Perhaps set an alarm on your phone as a reminder, and then put your phone in another room—out of sight, out of mind!

2.  **Warm Bath/Shower:**  Is a bath or shower more appealing to you?  Schedule it into your evening routine. Perhaps it could be right after your screen-free time begins.  Adding some Epsom salts or essential oils like lavender to your bath can enhance the relaxation factor even more.

3.  **Mindfulness/Meditation:** Start with short, guided meditations, even just 5-10 minutes. There are many free apps and online resources available.  Insight Timer, Calm, and Headspace are just a few examples.  Experiment and find one that you connect with.

So a sample routine could be:  Finish any necessary screen activities, set an alarm for 30 minutes before your target bedtime, then take your warm bath/shower. Once you are out, try a short guided meditation. Once you start to feel drowsy, head to bed. Continue this until it feels like a normal routine. How does this sound as a starting point? We can always adjust it as needed.

[USER]
That sounds like a great starting point. I think I can manage 30 minutes of screen-free time before bed, followed by a warm bath and a short meditation.  I prefer baths, so I'll try adding some Epsom salts or lavender oil for extra relaxation.

[COACH]
Wonderful! It’s great that you’re starting with what feels manageable and enjoyable.  Thirty minutes of screen-free time, a warm bath with Epsom salts or lavender oil, and a short meditation sounds like a truly relaxing pre-sleep ritual.  Remember, consistency is key.  Try to stick to this routine as much as possible, even on weekends, to help regulate your sleep-wake cycle.  

Let’s check in again next week.  How about we schedule a follow-up for then to see how the new routine is going, discuss any challenges you might be experiencing, and tweak things as needed?  We can also discuss how the alarm-free wake-up experiment is going and start figuring out that ideal bedtime based on your natural wake time.
\end{small}
\end{quote}

\end{document}